\documentclass[letterpaper, 10 pt, conference]{ieeeconf}

\IEEEoverridecommandlockouts
\makeatletter

\let\proof\@undefined
\let\endproof\@undefined
\makeatother

\usepackage{amsmath}  
\usepackage{amssymb}  
\usepackage{amsthm}
\usepackage{amsfonts}
\usepackage{mathtools}
\usepackage{paralist}
\usepackage{verbatim}
\usepackage{color}
\usepackage{graphicx}
\usepackage{times}

\usepackage{algorithm} 
\usepackage{algpseudocode}

\usepackage[us]{datetime}

\usepackage[dvipsnames,table]{xcolor}

\usepackage{soul}

\usepackage{todonotes}
\presetkeys{todonotes}{inline}{}

\usepackage{latexsym}
\usepackage{color}

\usepackage[noadjust]{cite}

\usepackage{float}
\usepackage{textcomp} 

\usepackage{tikz}
\usetikzlibrary{
  shapes,
  shapes.geometric,
  fit,
  arrows,
  calc,
  intersections,
  angles,
  positioning,
  backgrounds,
  decorations.markings,
  patterns,
  external
}

\tikzset{external/system call={latex \tikzexternalcheckshellescape -halt-on-error
-interaction=batchmode -jobname "\image" "\texsource";
dvips -o "\image".eps "\image".dvi;
ps2eps "\image.eps"}}

\let\labelindent\relax 
\usepackage[inline]{enumitem} 
\setenumerate[0]{label=\roman*.}

\usepackage{subfiles}
\usepackage{bm}
\usepackage[export]{adjustbox}
\usepackage{nicefrac}
\usepackage{subfig} 
\usepackage[compatibility=false]{caption}

\usepackage{siunitx}
\DeclareSIUnit \parsec {pc}
\DeclareSIUnit \electronvolt {eV}
\DeclareSIUnit \pixel {px}
\DeclareSIUnit \arcmin {arcmin}
\DeclareSIUnit \erg {erg}
\DeclareSIUnit \joul {J}

\tikzset{
  connect/.style args={(#1) to (#2) over (#3) by #4}{
    insert path={
      let \p1=($(#1)-(#3)$), \n1={veclen(\x1,\y1)},
      \n2={atan2(\x1,\y1)}, \n3={abs(#4)}, \n4={#4>0 ?180:-180}  in
      (#1) -- ($(#1)!\n1-\n3!(#3)$)
      arc (\n2:\n2+\n4:\n3) -- (#2)
    }
  },
  point/.style = {
    draw, circle,  fill = black, inner sep = 1.25pt
  },
}

\tikzset{
  base/.style = {
    rectangle,
    draw=black,
    fill=white,
    minimum height=0.7cm,
    text centered,
    font=\scriptsize,
    line width=0.2mm,
  },
  node/.style = {base, minimum width=1.65cm, line width=0.3mm},
  data/.style = {base, rounded corners=3mm, minimum width=1.65cm, line width=0.3mm},
  noborders/.style = {line width=0, font=\footnotesize},
}

\usepackage[nolist,nohyperlinks]{acronym}
\acrodef{GPS}[GPS]{Global Positioning System}
\acrodef{SLAM}[SLAM]{Simultaneous Localization And Mapping}
\acrodef{SLAMs}[SLAMs]{Simultaneous Localization And Mapping systems}
  \acrodef{RTK}[RTK]{Real-time Kinematics}
\acrodef{GNSS}[GNSS]{Global Navigation Satellite System}
\acrodef{ROS}[ROS]{Robot Operating System}
\acrodef{API}[API]{Application Programming Interface}
\acrodef{UAV}[UAV]{Unmanned Aerial Vehicle}
\acrodef{MAV}[MAV]{Micro Aerial Vehicle}
\acrodef{UGV}[UGV]{Unmanned Ground Vehicle}
\acrodef{UV}[UV]{Ultra-Violet}
\acrodef{LED}[LED]{Light-emitting Diode}
\acrodef{MBZIRC}[MBZIRC]{Mohamed Bin Zayed International Robotics Challenge}
\acrodef{DARPA}[DARPA]{Defense Advanced Research Projects Agency}
\acrodef{IMU}[IMU]{Inertial Measurement Unit}
\acrodef{LTI}[LTI]{Linear time-invariant}
\acrodef{MPC}[MPC]{Model Predictive Control}
\acrodef{UVDAR}[UVDAR]{Ultra-Violet Direction And Ranging}
\acrodef{DOF}[DOF]{degree-of-freedom}
\acrodef{DOFs}[DOFs]{degrees-of-freedom}
\acrodef{LiDAR}[LiDAR]{Light Detection and Ranging}
\acrodef{ESC}[ESC]{Electronic Speed Controller}
\acrodef{LKF}[LKF]{Linear Kalman Filter}
\acrodef{UKF}[UKF]{Unscented Kalman Filter}
\acrodef{EKF}[EKF]{Extended Kalman Filter}
\acrodef{RAS}[RAS]{Robotics and Automation Society}
\acrodef{IEEE}[IEEE]{Institute of Electrical and Electronics Engineers}
\acrodef{MRS}[MRS]{Multi-robot Systems Group}
\acrodef{FOV}[FOV]{Field of View}
\acrodef{CdTe}[CdTe]{Cadmium Telluride}
\acrodef{FDNPP}[FDNPP]{Fukushima Daiichi Nuclear Power Plant}

\newcommand{\reffig}[1]{Fig.~\ref{#1}}

\newcommand{\refsec}[1]{Sec.~\ref{#1}}
\newcommand{\reftab}[1]{Tab.~\ref{#1}}

\DeclareMathOperator*{\argmin}{\arg\!\min}

\makeatletter
\let\NAT@parse\undefined
\makeatother

\usepackage{hyperref} 
\hypersetup{
  colorlinks,
  citecolor=black,
  filecolor=black,
  linkcolor=black,
  urlcolor=black,
  pdfauthor={},
  pdfsubject={},
  pdftitle={}
}

\setlist[itemize]{noitemsep, nosep}

\usepackage{siunitx}
\usepackage{ifthen}

\usepackage{booktabs}

\renewcommand{\baselinestretch}{0.98}
\usepackage{tablefootnote}
\usepackage[hang,flushmargin]{footmisc}

\makeatletter
\def  \input@path{{./../fig/},{./fig/}}
\makeatother

\tikzset{
  imgletter/.style={
    rectangle,
    inner xsep=0.16em,
    inner ysep=0.18em,
    text=black,
    minimum height=0.8em,
    text centered,
    fill=white,
    fill opacity=0.8,
    text opacity=1,
    anchor=south west,
  },
  double arrow/.style args={#1 colored by #2 and #3}{
    -stealth,line width=#1,#2, 
    postaction={draw,-stealth,#3,line width=(#1)/3,
    shorten <=(#1)/3,shorten >=2*(#1)/3}, 
  }
}

\newcommand{\imagewithletter}[3]{
  \begin{tikzpicture}\label{#3}
      \node[anchor=south west,inner sep=0] (a) at (0,0) {
        #1
      };
      \begin{scope}[x={(a.south east)},y={(a.north west)}]
        \node[imgletter] (letter) at (0.0, 0.0) {\footnotesize{#2}};
        \draw (0.0, 0.0) rectangle (1.0, 1.0);
      \end{scope}
    \end{tikzpicture}
}

\usepackage{scalerel}
\usetikzlibrary{svg.path}
\definecolor{orcidlogocol}{HTML}{A6CE39}
\tikzset{
  orcidlogo/.pic={
    \fill[orcidlogocol] svg{M256,128c0,70.7-57.3,128-128,128C57.3,256,0,198.7,0,128C0,57.3,57.3,0,128,0C198.7,0,256,57.3,256,128z};
    \fill[white] svg{M86.3,186.2H70.9V79.1h15.4v48.4V186.2z}
    svg{M108.9,79.1h41.6c39.6,0,57,28.3,57,53.6c0,27.5-21.5,53.6-56.8,53.6h-41.8V79.1z M124.3,172.4h24.5c34.9,0,42.9-26.5,42.9-39.7c0-21.5-13.7-39.7-43.7-39.7h-23.7V172.4z}
    svg{M88.7,56.8c0,5.5-4.5,10.1-10.1,10.1c-5.6,0-10.1-4.6-10.1-10.1c0-5.6,4.5-10.1,10.1-10.1C84.2,46.7,88.7,51.3,88.7,56.8z};
  }
}
\newcommand\orcidicon[1]{\href{https://orcid.org/#1}{\mbox{\scalerel*{
  \begin{tikzpicture}[yscale=-1,transform shape]
    \pic{orcidlogo};
  \end{tikzpicture}
}{|}}}}

\title{
  RADRON: Cooperative Localization of Ionizing Radiation Sources by MAVs with Compton Cameras
}

\author{
  Petr Stibinger$^{1}$$^{\orcidicon{0000-0002-7662-9230}}$,
  Tomas Baca$^{\phantom{.}}$$^{\orcidicon{0000-0001-9649-8277}}$,
  Daniela Doubravova$^{\phantom{.}}$$^{\orcidicon{}}$,
  Jan Rusnak$^{\phantom{.}}$$^{\orcidicon{}}$,\\
  Jaroslav Solc$^{\phantom{.}}$$^{\orcidicon{}}$,
  Jan Jakubek$^{\phantom{.}}$$^{\orcidicon{0000-0003-0661-8584}}$,
  Petr Stepan$^{\phantom{.}}$$^{\orcidicon{0000-0002-7444-3264}}$
  and Martin Saska$^{\phantom{.}}$$^{\orcidicon{0000-0001-7106-3816}}$
}

\begin{document}

\maketitle

\minipage[c][0.1\textheight][s]{\textwidth}
\vspace{-5cm}
\footnotesize
  This work has been submitted to the IEEE for possible publication. Copyright may be transferred without notice, after which this version may no longer be accessible.
\endminipage



\vspace{-2.31cm}
\begin{abstract}
  We present a novel approach to localizing radioactive material by cooperating \acp{MAV}.
  Our approach utilizes a state-of-the-art single-detector Compton camera as a highly sensitive, yet miniature detector of ionizing radiation.
  The detector's exceptionally low weight (\SI{40}{\gram}) opens up new possibilities of radiation detection by a team of cooperating agile \acp{MAV}.
  We propose a new fundamental concept of fusing the Compton camera measurements to estimate the position of the radiation source in real time even from extremely sparse measurements.
  The data readout and processing are performed directly onboard and the results are used in a dynamic feedback to drive the motion of the vehicles.
  The \acp{MAV} are stabilized in a tightly cooperating swarm to maximize the information gained by the Compton cameras, rapidly locate the radiation source, and even track a moving radiation source.
\end{abstract}






\section{INTRODUCTION}


Nuclear environments represent a domain particularly well suited for the deployment of mobile robots \cite{sanada2015aerial, jiang2016prototype, mascarich2018radiation}.
The primary driving force is to reduce human exposure to harmful radiation, and to facilitate access to areas that are difficult to reach by conventional means.
\acp{UAV} provide the capability to swiftly cover expansive areas, making them ideal for rapid response as well as repeated surveillance.
Large aerial platforms offer higher payload capacity and operational endurance,
while compact \acp{MAV} can operate much closer to the radiation sources, where the flux of ionizing particles is significantly denser.
The reduced size and cost of smaller platforms open up a potential for multi-robot deployment, where the \acp{MAV} can operate in compact groups and act as a dynamic sensor network.

Nuclear environments can emerge suddenly---as a result of natural disasters, accidents, or criminal activities---or develop gradually through heavy industrial activity.
In all cases, the terrain conditions are expected to be rugged and obstructed by debris, ongoing cleanup efforts, or nature reclaiming the depopulated areas.
Aerial vehicles hold a clear advantage over ground-based platforms in rapid traversal of the area regardless of terrain conditions.

\begin{figure}[!t]
  \centering
  \resizebox{\columnwidth}{!}{
    \input{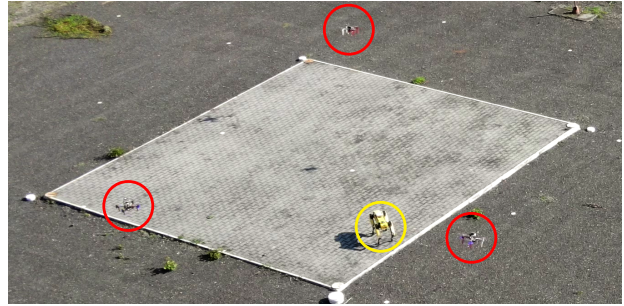}
  }
  \caption{
    Three cooperating \acp{MAV} localizing and tracking a moving radiation source.
    Each \ac{MAV} (red) is equipped with a miniature Compton camera and an onboard computer.
    The ionizing radiation source (a sample of Cesium-137) is carried by a quadruped robot (yellow).
  }
  \label{fig:intro_fig}
\end{figure}

Continuous developments in radiation sensing technology have enabled the construction of extremely sensitive detectors, which are capable of distinguishing individual high-energy photons.
This gave rise to a new type of sensors---Compton cameras---which utilize the predictable properties of Compton scattering \cite{compton1923quantum} to determine a set of possible directions towards the radiation source.
Compton cameras were pioneered in high-energy astronomy, and originally consisted of bulky scintillators and photomultipliers, which are impractical for use in aerial robotics.
However, compact Compton cameras are slowly becoming available \cite{parajuli2022development}, comprising of miniaturized scintillators \cite{jiang2016prototype, sato2018radiation, sato2019radiation} or semiconductor chips using direct energy conversion \cite{watanabe2007development, turecek2018compton}.

A conventional Compton camera consists of two stacked detectors---a scatterer and an absorber.
Such a configuration requires additional electronics for synchronization, and necessitates active cooling, making it impractical for use onboard an \ac{MAV}.
In this work, we exploit the cutting-edge advances in the semiconductor technology, which made a single-detector Compton camera possible.
The reduced complexity makes this type of detector a great fit for sub-\SI{3}{\kilo\gram} \acp{MAV}.



\section{RELATED WORK}
The area of the \ac{FDNPP} has been inspected by several robotic missions since the disaster in 2011.
Initial surveys were performed by \acp{UAV} equipped with powerful but heavy scintillator detectors \cite{towler2012radiation, sanada2015aerial, jiang2016prototype, christie2017radiation}.
Similar missions were also conducted to monitor the environmental impact of uranium ore extraction at several mining sites \cite{martin2015use, salek2018mapping, keatley2018source, kunze2022development}.
In the aforementioned works, the authors rely on large \acp{UAV} which are only suitable for open areas where a \ac{GNSS} is available.
However, recent advances in onboard sensing and \ac{GNSS}-denied localization \cite{baca2021mrs} have made it possible to deploy \acp{MAV} directly inside mine shafts \cite{roucek2019darpa}, cave systems \cite{petrlik2020robust}, and building interiors \cite{kratky2020autonomous, saska2020formation, pritzl2021icra} including nuclear power plants \cite{kratky2020autonomous}.

The research on ionizing radiation localization is currently split into two major streams: the localization of discrete radiation sources, and the identification of a continuous contamination by aerosols or dissolved nuclear material.
The problem of discrete radiation source localization may be tackled by adapting well-known robotic principles such as maximum likelihood estimation
\cite{cordone2017improved}, gradient descent \cite{baidoo2013gradient}, Kalman filters \cite{baca2021gamma}, and particle filters \cite{gao2018robust, pinkam2020informative, anderson2020mobile}.
For spatially distributed radioactive fields, the aim is to estimate the spatial representation of the radiation field and its volumetric properties.
Related research has demonstrated feasible approaches using a Gaussian mixture model \cite{morelande2009radiation}, maximum likelihood estimation \cite{kim20173d}, contour analysis \cite{towler2012radiation, newaz2016uav}, and gradient descent \cite{mascarich2019distributed, mascarich2021autonomous}.

The localization approaches often rely on teleoperation \cite{aleotti2017detection}, or following pre-planned waypoints \cite{macfarlane2014lightweight, martin20163d, sanada2015aerial, han2013low, kochersberger2014post}.
An active approach, utilizing a grid-based Bayesian estimator in a direct motion planning feedback loop, is presented in \cite{towler2012radiation} using a \SI{90}{\kilo\gram} \ac{UAV}.
The approach in \cite{christie2017radiation} employs a cooperative approach, using a \SI{94}{\kilo\gram} \ac{UAV} following a space-filling path to map the area of interest and to identify radiation hotspots for further inspection by a companion \ac{UGV}.
Similar work was presented in \cite{kochersberger2018unmanned}, where a \SI{16.8}{\kilo\gram} \ac{UAV} is used for exhaustive aerial mapping, and an autonomous companion \ac{UGV} is later deployed for a close-up hotspot inspection.
In \cite{schraml2022real} a large eight-rotor \ac{UAV} capable of carrying up to \SI{16}{\kilo\gram} of payload is used.
The authors use a heavy gamma spectrometer, enabling the localization of multiple radiation sources in a semi-autonomous mode.
In \cite{mascarich2018radiation}, a small \SI{2.6}{\kilo\gram} six-rotor \ac{MAV} equipped with a scintillation detector employs an active radiation search strategy, using information gain and distance from the radiation field centroid in further path-planning.
The authors of \cite{mascarich2023autonomous} demonstrate an informative path planning and characterization of a distributed radiation field by a single \ac{MAV} weighing just \SI{639}{\gram}.
Contrary to our approach, both the platform and methodology in \cite{mascarich2023autonomous} are heavily optimized for indoor deployment and an operational time $<$ \SI{6}{\minute}.
In \cite{wernerComptonRadiation2024}, a method for localizing multiple sources of radiation by a team of \acp{MAV} is introduced.
Nonetheless, their approach treats the \acp{MAV} as independent agents operating in a shared map, thus not fully exploiting the information from concurrent measurements taken at different locations.

The simulations in \cite{baca2021gamma, gu2020uav} demonstrate, that a single \ac{MAV} may be used to follow a moving radiation source under certain conditions.
However, there is a consensus that multiple networked sensors are required to accurately track a moving radiation source \cite{liu2011sensor, ahmad2021ionizing, victor2015mobile}.
To the best of our knowledge, cooperative localization and tracking of a moving ionizing radiation source has not been tackled in the existing literature.
This work distinguishes itself from the state of the art by leveraging a team of networked, tightly cooperating \acp{MAV} to localize and track the radiation source.


\section{CONTRIBUTION}
We contribute a decentralized multi-robot approach for localizing an ionizing radiation source that diverges from the conventional approaches relying on a single, heavy platform.
Our solution uniquely combines the compact, extremely lightweight MiniPIX TPX3 Compton camera with a self-organized swarm of tightly cooperating \acp{MAV}, enabling deployment in environments previously inaccessible to systems relying on bulky detectors and robots.
We extend the Compton data fusion method from \cite{baca2021gamma} to the multi-robot domain and introduce a novel control law that uses distributed radiation measurements processed in real time to dynamically drive the motion of the swarm.
The problem represents a rare instance where the use of a tightly cooperating self-organized swarm is fundamentally justified, as our unique approach fully utilizes the availability of distributed radiation measurements processed directly onboard.
This removes the need to perform long exposition imaging, which would be otherwise necessary with miniaturized radiation sensors, and results in a key advantage---the ability to localize and track a moving radiation source in real time.
This capability is not attainable, or is severely limited with the state-of-the-art single-vehicle methods.
We empirically validate all stated capabilities in realistic simulations, as well as real-world experiments.



\section{MEASUREMENT AND DATA FUSION}
\label{sec:measurement_and_data_fusion}
\subsection{Single-detector Compton camera}
The MiniPIX TPX3 Compton camera leverages the predictable mechanics of Compton scattering to estimate the direction of incoming ionizing radiation.
Compton scattering occurs when a high-energy photon interacts with a charged particle, typically an electron.
During this interaction, the photon transfers a portion of its energy to the charged particle, and scatters in a new direction.
The energy of the scattered photon $E_{\lambda'}$ is derived from \cite{compton1923quantum} as:
\begin{equation}
  \phantom{.,}E_{\lambda'} = \frac{E_\lambda}{1 + (E_\lambda/(m_ec^2))(1-\mathrm{cos}\phantom{.}\theta)}\phantom{.},
\end{equation}
where $E_\lambda$ is the initial photon energy, $m_e$ is the rest mass of an electron, $c$ is the speed of light in vacuum, and $\theta$ is the scattering angle.
The initial energy $E_\lambda$ is unknown, however, the conservation of energy for the scattering event holds:
\begin{equation}
  \phantom{,.}E_\lambda = E_{\lambda'} + E_{e'}\phantom{.},
\end{equation}
where $E_{e'}$ is the energy of the recoiled electron.
If the recoiled electron and scattered photon are detected, it is possible to reconstruct the scattering angle $\theta$ as:
\begin{equation}
  \phantom{..}\theta = \mathrm{arccos}\left(1 + m_ec^2\left(\frac{1}{E_{\lambda}}-\frac{1}{E_{\lambda'}}\right)\right)\phantom{.}.
  \label{eqn:compton}
\end{equation}

The Timepix3 chip inside the camera consists of a $256 \times 256$ pixel grid on a \SI{14}{\milli\meter}$\times$\SI{14}{\milli\meter} CdTe chip.
Each pixel operates as an independent radiation detector capable of direct energy conversion.
The 2D position of both scattering products can be accurately determined from the indices of activated pixels.
Additionally, it is possible to infer the relative distance between the scattering products from the difference in detection times and the velocity of signal propagation through the detector.
The velocity is obtained from a calibration process with nanosecond precision \cite{baca2021gamma, turecek2020single}.

\begin{figure}[htbp]
  \centering
  \includegraphics[width=1.0\columnwidth]{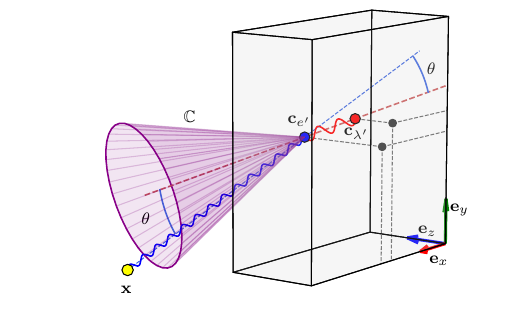}
  \caption{
    Cone reconstruction in a single-detector Compton camera.
    A high-energy photon (blue) is emitted by a source $\mathbf{x}$.
    The Compton scattering occurs on the first contact with the detector and the recoiled electron is immediately absorbed.
    The scattered photon (red) passes through the detector until it is also absorbed.
    Energies $E_{\lambda'}$, $E_{e'}$ are measured directly by the pixel matrix bonded to the rear face of the detector.
    The active pixels pinpoint the $x, y$ coordinates of both events $\mathbf{c}_{e'}, \mathbf{c}_{\lambda'}$.
    The difference in $z$ is computed from the time delay between the two detection events and the velocity of event propagation (obtained in calibration).
    The scattering angle $\theta$ is computed using (\ref{eqn:compton}).
    Finally, a set of all possible origins of the high-energy photon is reconstructed.
    This set forms the surface of a cone $\mathbb{C}$.
  }
  \label{fig:compton_reconstruction}
\end{figure}

As the third coordinate cannot be measured in absolute terms, an assumption is made, that the Compton scattering occurs directly on the outer edge of the detector.
This yields the 3D coordinates $\mathbf{c}_{e'}, \mathbf{c}_{\lambda'}$ for the recoiled electron and the scattered photon, respectively.
Finally, the Compton cone $\mathbb{C}$ is reconstructed (\reffig{fig:compton_reconstruction}).
The cone axis is given by $\mathbf{c}_{e'}, \mathbf{c}_{\lambda'}$, the apex angle $\theta$ is given by (\ref{eqn:compton}), and the apex point coincides with $\mathbf{c}_{e'}$.
The surface of the cone represents a set of possible origins of the incoming ionizing radiation.
The assumption made here does not influence the shape of the cone, it only shifts it along the detector's $z$-axis.
The detector thickness in this direction is only \SI{2}{\milli\meter}, therefore the error introduced is negligible when considering the real-world distances involved in target localization by \acp{MAV}.

\subsection{Data fusion for multiple Compton cameras}
The position of a radiation source is conventionally estimated by projecting the Compton cones into a simple geometric approximation of the environment, such as a 2D projection plane \cite{sato2019radiation, turecek2020single}, a sphere centered around the detector \cite{haefner2015filtered}, or a 3D voxel grid \cite{kim20173d}.
However, such approaches require long-exposition imaging with a stationary Compton camera.
This strategy is not feasible for compact \acp{MAV} with a limited operational time.
Instead, we exploit the mobility of the \acp{MAV} and employ a novel approach to Compton cone fusion, which enables rapid estimation from only a few samples.
The approach relies on event-based processing of individual Compton scattering occurrences while the \ac{MAV} is moving through the environment at high speed.

\begin{figure}[htbp]
  \centering
  \includegraphics[clip, trim=0 1.1cm 0 0.7cm, width=1.0\columnwidth]{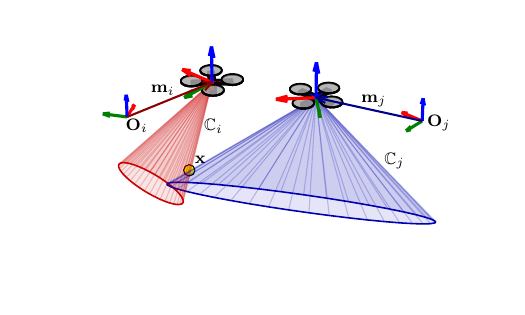}
  \caption{
    An example with 2 \acp{MAV} using heterogeneous positioning systems. Each drone $i,j$ estimates its position and orientation relative to the origin $\mathbf{O}_i, \mathbf{O}_j$, respectively. The Compton cones $\mathbb{C}_i, \mathbb{C}_j$ are also projected in the respective coordinate frames. Under (A2), the position of the radiation source $\mathbf{x}$ can be estimated in both coordinate frames using all available measurements.
  }
  \label{fig:multiple_projections}
\end{figure}

Our data fusion method relies on the \ac{LKF}, which is well suited for single-target localization.
Additionally, the algorithm is computationally inexpensive, which is crucial for real-time onboard estimation performed by a resource-constrained \ac{MAV}.
Due to the limited information provided by isolated measurements, we do not model the dynamics of the target in the estimation process, and instead utilize the track-by-detection approach.
Therefore, the state-space model of the radiation source is reduced to:
\begin{equation}
  \phantom{,.}\mathbf{x}_k = \mathbf{F}\mathbf{x}_{k-1}\phantom{.},
\end{equation}
\noindent where $\mathbf{x}_k=[x_x\text{ }x_y\text{ }x_z]^T$ is the estimated position of the radiation source at step $k$, and $\mathbf{F}$ is an identity matrix $\mathbf{I}_{3\times3}$.
To input a new measurement to the \ac{LKF}, the Compton cone $\mathbb{C}$ has to be transformed into a vector, as closer described in \cite{baca2021gamma}.
Here, we generalize the process so that an arbitrary number of detectors and robots can be used simultaneously.
In the following text, we refer to the position of the radiation source estimated by the \ac{LKF} as a \emph{hypothesis}.
The generalization is done under the following assumptions:

\begin{enumerate}[label=(A\arabic*), ref=(A\arabic*)]
  \item \ac{MAV} $i$ estimates its own position and orientation in the environment as $\mathbf{m}_i$ relative to the origin $\mathbf{O}_i$. \label{assumption1}
  \item A transformation $\mathbf{T}_{i,j}$ exists between the coordinate frames used by \acp{MAV} $i,j$, s.t. $\mathbf{m}_j = \mathbf{T}_{i,j}\mathbf{m}_i$. \label{assumption2}
  \item The optimal measurement correction places the current hypothesis $\mathbf{x}_k$ to the nearest point on the surface of the cone $\mathbb{C}_k$. \label{assumption3}
\end{enumerate}

The assumptions \ref{assumption1}, \ref{assumption2} permit the use of heterogeneous positioning systems within the swarm, as illustrated in \reffig{fig:multiple_projections}.
Under these assumptions, all \acp{MAV} are able to utilize the full set of measurements collected by the swarm, and perform the data fusion and motion planning directly onboard.

\begin{figure}[htbp]
  \centering
  \includegraphics[clip, trim=0 0.3cm 0 0.2cm, width=0.95\columnwidth]{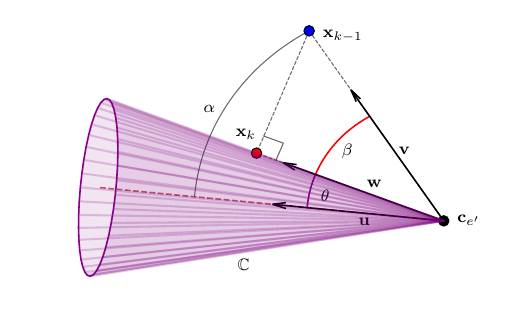}
  \caption{
    Measurement correction done by projecting the latest hypothesis $\mathbf{x}_{k-1}$ onto the surface of a cone $\mathbb{C}$ to get a corrected hypothesis $\mathbf{x}_k$.
  }
  \label{fig:orthogonal_projection}
\end{figure}

The geometric representation of \ref{assumption3} depends on the orientation of $\mathbb{C}$, and is either the apex of the cone, or an orthogonal projection of the previous hypothesis $\mathbf{x}_{k-1}$ onto the surface of the cone (\ref{eqn:projection}).
To compute the orthogonal projection (\reffig{fig:orthogonal_projection}), let $\mathbf{u}$, $\mathbf{v}$ be unit vectors originating at the cone's apex $\mathbf{c}_{e'}$.
Vector $\mathbf{u}$ is collinear with the axis of the cone and $\mathbf{v}$ is oriented towards $\mathbf{x}_{k-1}$.
These two vectors unambiguously define a plane in 3D.
The angle $\alpha$ between $\mathbf{u}$ and $\mathbf{v}$ is then given as:
\begin{equation}
  \phantom{.,}\alpha = \mathrm{arccos}\left(\mathbf{u}\cdot\mathbf{v}\right) = \theta + \beta\phantom{.},
\end{equation}
\noindent where $\theta$ is the scattering angle and $\beta$ is the angle between the surface of cone $\mathbb{C}$ and $\mathbf{v}$.
Rotating $\mathbf{v}$ by the angle $-\beta$ yields vector $\mathbf{w}$.
The hypothesis $\mathbf{x}_k$ is projected orthogonally on the surface of $\mathbb{C}$.
Finally, $\mathbf{x}_k$ forms a right-angled triangle with $\mathbf{x}_{k-1}$ and $\mathbf{c}_{e'}$, and $\mathbf{x}_k$ is then computed as:
\begin{equation}
  \phantom{.,}\mathbf{x}_k =
  \begin{cases}
    \mathbf{c}_{e'} + \mathbf{w} \cdot \lVert \mathbf{x}_{k-1} -  \mathbf{c}_{e'} \rVert \cdot \mathrm{cos}\phantom{.}\beta, & \mathrm{if}\text{ }\alpha < \pi/2\phantom{.},\\
    \mathbf{c}_{e'}, & \mathrm{otherwise}.
  \end{cases}
  \label{eqn:projection}
\end{equation}

Since the orthogonal projection moves the hypothesis in a known direction, we model the measurement covariance $\mathbf{R}$ accordingly in each correction step of the \ac{LKF}.
A canonical covariance matrix $\mathbf{R_C}$ is created with the measurement covariance $\rho$ assigned to the first coordinate.
The other elements on the main diagonal are assigned values orders of magnitude larger than $\rho$.
A rotation $\mathbf{P}$ is then applied, s.t. the X-axis is collinear with the projection axis ($\mathbf{x}_k - \mathbf{x}_{k-1}$).
The result is a covariance matrix $\mathbf{R}$ that only increases the estimate confidence along the axis of orthogonal projection:
\begin{equation}
  \phantom{..}\mathbf{R_C} = \left(\begin{smallmatrix}
    \rho & 0 & 0 \\
    0 & \rho\cdot10^4 & 0 \\
    0 & 0 & \rho\cdot10^4
  \end{smallmatrix}\right),\phantom{.}
  \mathbf{R} = \mathbf{PR_CP}^T\phantom{.}.
\end{equation}

\section{COOPERATIVE LOCALIZATION STRATEGY}
\label{sec:cooperative_localization_strategy}

This section outlines the novel cooperative radiation source localization strategy using a swarm of \acp{MAV}.
The mission is divided into two distinct stages: an initial systematic search aimed at quickly acquiring an approximate position of the source, and a cooperative tracking phase where the \ac{MAV} swarm dynamically centers itself on the hypothesis and continuously refines the estimate.
A key aspect is the decentralized control and estimation pipeline, where each \ac{MAV} performs its own data fusion and motion planning onboard, enabled by sharing the Compton measurements and \ac{MAV} positions among all swarm members.

\subsection{Hypothesis initialization}
The initial stage focuses on rapidly acquiring the initial hypothesis $\mathbf{x}_0$ to initialize the \ac{LKF}.
The data fusion approach described in \refsec{sec:measurement_and_data_fusion} relies on recursive projection of the previous hypothesis onto the surface of a new Compton cone.
Since individual Compton measurements do not convey any information about the distance between the detector and the radiation source, the initialization has to be performed by gathering $M$ measurements taken at different locations.

The area of interest is divided into $N$ equal segments, and a space-filling path is generated for each segment.
Each \ac{MAV} is then assigned one path to explore one segment of the area.
While the systematic search is ongoing, the \acp{MAV} maintain a fixed velocity and continuously change their heading to minimize potential blind spots of the forward-mounted Compton cameras.
During this stage, the \acp{MAV} operate independently, but share all Compton measurements with the other vehicles.

Once the cumulative amount of Compton cones reaches $M$, the initial hypothesis $\mathbf{x}_0$ is computed.
This computation is formulated as a non-linear least squares optimization task to find a point that minimizes the squared distance to all the surfaces of the $M$ cones.
The formulation of the optimization task is detailed in \cite{baca2021gamma}.
Solving the optimization task is computationally intensive, therefore, it is only performed once to acquire $\mathbf{x}_0$.


\subsection{Cooperative source localization and tracking}
Upon acquiring $\mathbf{x}_0$, the mission enters the second stage where all \acp{MAV} operate as a tightly cooperating, self-organized swarm.
At this stage, the goal is to maximize the information gained by the miniature Compton cameras and to continuously update the hypothesis, enabling the swarm to localize and track the radiation source.

A single Compton camera event yields a set of possible directions towards the radiation source.
This set can be narrowed down by consecutive measurements.
However, even a long-exposition imaging by a stationary camera capturing thousands of events, will only yield a direction vector towards the radiation source.
Moreover, the estimation may converge at a local optimum located at the position of the Compton camera itself, as all cones have their apex points concentrated on the miniature surface of the detector.
Therefore, it is necessary to keep the cameras in motion, so that subsequent measurements are taken at different points in the environment.

Following a circular path centered at the hypothesis ensures, that the camera moves orthogonally to the direction to the hypothesis, and the baseline between consecutive measurements is maximized.
Using more than one \ac{MAV} increases the overall sensor volume and inherently provides measurements from multiple viewpoints.
This dramatically enhances the sensitivity and the ability to track a moving radiation source.
The simultaneous use of multiple \acp{MAV} necessitates a high-level controller to coordinate motion and efficiently distribute the Compton cameras.
We implement a novel decentralized flocking control leveraging the computationally inexpensive data fusion approach introduced in \refsec{sec:measurement_and_data_fusion}.
Each swarm member performs the fusion directly onboard, using the results in feedback for motion planning.


\subsection{Compton-driven flocking}

\begin{table}[htbp]
  \centering
  \caption{
    Notation used in the flocking algorithm description
    \label{tab:swarm_notation}
  }
  \begin{tabular}{l l}
    \toprule
    $N$ & number of \acp{MAV} in the swarm \\
    $r$ & desired circle radius \\
    $v$ & desired tangential speed\\
    $K$ & number of trajectory generation steps \\
    $\mathbf{t}_i\left[k\right]$ & $k$-th trajectory step of the $i$-th \ac{MAV} \\
    $\mathbf{x}$ & estimated position of the radiation source (hypothesis) \\
    $\mathbf{m}_i=\left(r_i\text{ }\varphi_i\right)^T$ & position of the $i$-th \ac{MAV} in polar coordinates \\
    $\theta_i$ & smallest angle from the $i$-th \ac{MAV} to another \ac{MAV}\\
    $\theta^*$ & uniform spacing angle \\
    \bottomrule
  \end{tabular}
\end{table}

The flocking controller aims to distribute the \acp{MAV} on a circle around the hypothesis while maintaining a desired radius $r$ and tangential speed $v$.
The core objective is to achieve a uniform spacing angle $\theta^*=(2\pi/N)$~rad between neighboring \acp{MAV} to maximize the baseline between the Compton cameras.

The Compton cones reconstructed from individual measurements, along with the current position estimates, are shared between the \acp{MAV}.
Under assumption \ref{assumption2}, each \ac{MAV} can utilize this data within its own frame of reference.
This ensures all \acp{MAV} operate with the same hypothesis and plan their own path on the same circle.

The radioactive object is assumed to be a point source located on the surface of the environment, with a uniform spatial distribution of radiation.
The flocking algorithm therefore focuses on the horizontal motion of the \acp{MAV}, assuming they all operate at the same height above ground.
We describe the problem in the plane of encirclement using polar coordinates with the hypothesis $\mathbf{x}$ placed at the origin.
We utilize a trajectory generation approach rather than a traditional feedback controller, due to the step-like disturbances introduced by the irregular updates of the hypothesis by the \ac{LKF}, which occur with each new Compton cone reconstruction.
Each swarm member is assigned a unique ID $i\in\{1,\dots,N\}$, and the positions of the \acp{MAV} are $\mathbf{m}_i=\left[r_i\text{, }\varphi_i\right]^T$.

The trajectory is generated for a single-vehicle as a circular arc with radius $r$.
The arc is divided into $K$ samples using an arbitrary sampling period $\Delta T$ to achieve a desired tangential speed $v$.
To enforce uniform distribution of the \acp{MAV} on a circle, the trajectories are offset using a bias angle $\beta_i$.
Firstly, the angle $\theta_i$ is computed as the smallest oriented angle between the $i$-th \ac{MAV} and its neighbors:
\begin{equation}
  \phantom{,}\theta_i = \mathrm{AngleDiff}\left(\mathbf{m}_i, \mathbf{m}_j\right),
\end{equation}
\noindent where $j$ is the index of the $i$-th \ac{MAV}'s nearest neighbor in terms of the central angle $\varphi$:
\begin{equation}
  \phantom{.}j = \argmin_{j \in N\smallsetminus i}\left|\mathrm{AngleDiff}\left(\mathbf{m}_i, \mathbf{m}_j\right)\right|.\\
\end{equation}
\noindent The bias $\beta_i$ is then applied in the direction opposite to $\theta_i$, acting as a repulsive force between the swarm members:
\begin{equation}
  \phantom{..}\beta_i =
    -\mathrm{sgn}\left(\theta_i\right)\beta\phantom{.}.
\end{equation}
\noindent Finally, the desired trajectory $\mathbf{t}_i$ is generated as a sequence:
\begin{equation}
  \phantom{..}\mathbf{t}_i[k] = \left[r\text{, }\varphi_i + \beta_i + krv\Delta T\right]^T\phantom{.}, \forall k \in \left\{0,\dots,K\right\}.
  \label{eqn:flocking_trajectory_generation}
\end{equation}

\begin{figure}[htbp]
  \centering
  \includegraphics{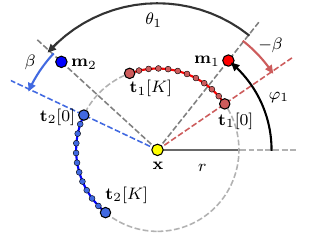}
  \caption{
    A top-down view of the swarm's self-organization during the encirclement of the hypothesis.
    Here shown for two \acp{MAV} with starting positions $\mathbf{m}_{1}, \mathbf{m}_{2}$.
    The signed angle from $\mathbf{m}_{1}$ to the nearest other \ac{MAV} is $\theta_1$.
    To steer the swarm towards a uniform spacing angle $\theta^{*}$, a bias $\beta$ is applied to the start of the generated trajectory $\mathbf{t}_1$.
    The bias is applied in the orientation opposite to $\theta$.
    The first point of the desired trajectory $\mathbf{t}_1[0]$ does not correspond to the position of the \ac{MAV} $\mathbf{m}_1$.
    However, the discontinuities in the desired trajectory are already tackled at a lower level of the control architecture \cite{baca2018model}.
    This allows the swarm controller to retain simplicity and generate new trajectories at a high rate.
  }
  \label{fig:swarm_control}
\end{figure}

Without loss of generality, the process of trajectory generation can be illustrated for a 2-vehicle case (\reffig{fig:swarm_control}).
It is worth noting, that the initial point of the trajectory $\mathbf{t}_i[0]$ may not coincide with the current position of the \ac{MAV} $\mathbf{m}_i$.
The discontinuity is tackled by passing the desired trajectory through the MPC trajectory tracker \cite{baca2018model} further in the control pipeline.
The tracker then outputs a feasible tracking reference for the \ac{MAV}'s flight controller.
The resulting control reference satisfies the vehicle's velocity and acceleration constraints and is optimal with respect to the tracker's dynamic model.
The complete flow of data within one \ac{MAV} is shown in \reffig{fig:data_flow_diagram}.

\begin{figure}[htbp]
  \centering
    \begin{tikzpicture}[>=latex, node distance=0.33cm and 0.33cm]
  \newcommand{\rows}[2]{\begin{tabular}{c} #1 \\ #2 \end{tabular}}

    \node (compton) [node] {\rows{Compton}{camera}};
    \node (fusion) [node, right= of compton] {\rows{Cone}{fusion}};
    \node (flocking) [node, right= of fusion] {\rows{Flocking}{control}};
    \node (tracker) [node, right= of flocking] {\rows{MPC}{tracker}};
    
    \node (cones) [data, below= of compton, xshift=0.5cm]{\rows{Compton}{cones}};
    \node (estimate) [data, below= of fusion, xshift=0.5cm] {\rows{Source}{estimate}};
    \node (trajectory) [data, below= of flocking, xshift=0.5cm] {\rows{Desired}{trajectory}};
    \node (reference) [data, below= of tracker, xshift=0.5cm] {\rows{Control}{reference}};

    \begin{pgfonlayer}{background}
      \draw[->] ([xshift=0.2cm]compton) |- (cones) -| ([xshift=-1.5cm]fusion);
      \draw[->] ([xshift=0.2cm]fusion) |- (estimate) -| ([xshift=-1.5cm]flocking);
      \draw[->] ([xshift=0.2cm]flocking) |- (trajectory) -| ([xshift=-1.5cm]tracker);
      \draw[->] ([xshift=0.2cm]flocking) |- (trajectory) -| ([xshift=-1.5cm]tracker);
      \draw[->] ([xshift=0.2cm]tracker) |- (reference) -- ([xshift=0.3cm]reference.east);
    \end{pgfonlayer}

\end{tikzpicture}
  \caption{
    A data flow diagram for one of the \acp{MAV}.
    The decentralized flocking control is driven by the fusion of radiation measurements from all swarm members performed onboard each vehicle in real time.
    This allows the system to rapidly react to sudden changes, such as the movement of the radiation source.
    The desired trajectory is passed through the MPC trajectory tracker \cite{baca2018model} which ensures the control reference is feasible with respect to the dynamic constraints of the vehicle.
}
  \label{fig:data_flow_diagram}
\end{figure}



\section{SIMULATIONS}

This section presents a series simulations designed to rigorously evaluate the novel cooperative radiation localization strategy introduced in \refsec{sec:cooperative_localization_strategy}.
The primary goal is to validate the capabilities of the proposed system and to highlight the advantages of using a swarm over a single \ac{MAV}.
Additionally, we focus on tuning the control parameters, notably the encirclement radius $r$, tangential speed $v$, and swarm size $N$, to outline a suitable configuration for real-world deployment.

\subsection{Swarm stabilization}
\begin{figure}[htbp]
  \centering
    \includegraphics{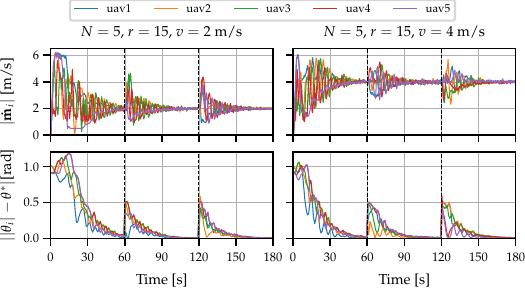}
  \caption{
    Simulation results showing the swarm stabilizing after a step disturbance.
    The flocking algorithm is initialized with 5 \acp{MAV} in arbitrary positions.
    The hypothesis is static and every \SI{60} seconds its position is shifted, forcing the swarm to reorganize around a new center.
    The upper row shows the instantaneous speed $|\mathbf{\dot{m}}_i|$ of the \acp{MAV}.
    The lower row shows the magnitude of difference between the uniform spacing angle $\theta^*$ and the relative angle $\theta_i$ to the nearest neighbor of each \acp{MAV}.
    The speed of all \acp{MAV} converges at the desired speed $v$, and the angular spacing converges to a uniform distribution.
}
  \label{fig:swarm_stabilization}
\end{figure}

A series of simulations was designed to validate that the proposed flocking controller (\ref{eqn:flocking_trajectory_generation}) is capable of stabilizing the swarm from an unorganized initial state and recover after a step disturbance.
We decouple the Compton cone fusion from the motion planning, and use a static hypothesis in this simulation series.
The \acp{MAV} are initially placed at an arbitrary position and the flocking controller is engaged.
A disturbance is introduced by changing the position of the hypothesis by \SI{10}{\meter} with a \SI{60}{\second} period.
The disturbance shifts the center of the circular trajectory and forces the swarm to reorganize.
The results (\reffig{fig:swarm_stabilization}) indicate that the flocking controller is able to stabilize the swarm in the circular motion, achieve the desired speed $v$, and uniformly distribute the \acp{MAV} around the hypothesis.

\subsection{Radiation source localization}

Building on the capabilities of a self-stabilizing \ac{MAV} swarm, this section tackles the core objective: the cooperative radiation source localization.
Through a series of realistic simulations, we validate the system's effectiveness in localization and tracking of a moving radiation source, and highlight the key advantages over the state-of-the-art single-vehicle methods.
The simulations are performed in a $100\times100$~m area.
A cooperative approach using a 3 \acp{MAV} is compared to the single-\ac{MAV} approach using the same control parameters: $r=12\phantom{.}$\SI{}{\meter}, $v=3\phantom{.}$\SI{}{\meter\per\second}, $K=30$.
The radiation source is Cesium-137 with activity \SI{3}{\giga\becquerel}, and for the tracking experiments it follows a circular path with a \SI{40}{\meter} radius at a speed of \SI{1}{\meter\per\second}.
We focus on the comparative qualities of both mission stages: hypothesis initialization and radiation source tracking.
The simulation is terminated when a Compton cone is not received for more than \SI{20}{\second}, indicating a loss of the target, or after \SI{180}{\second} of continuous tracking.
The simulation is repeated 50 times for each configuration, with randomized initial position of the radiation source, to enable both quantitative and qualitative evaluation.
The simulation results are summarized in \reftab{tab:simulation_results} and clearly highlight the comparative advantages of the \acp{MAV} swarm over a single \ac{MAV} in localizing both static and moving radiation sources.
With the novel swarm approach, the time to acquire the initial hypothesis is dramatically reduced to just 25\% of the single-vehicle requirement.
It is also worth noting, that the median estimation error of a single vehicle reaches over \SI{20}{\meter} when the radiation source is moving, which significantly limits any practical use of this approach.
The use of a swarm, on the other hand, improves the accuracy of tracking a moving source by nearly 300\%, and the median error achieved by a swarm of just 3 \acp{MAV} would be sufficient to pinpoint a specific vehicle.

\begin{table}[htbp]
  \centering
  \caption{
    \label{tab:simulation_results}
    Localization and tracking simulation results
  }
  \begin{tabular}{l cccc}
    \toprule
    & \multicolumn{2}{c}{\textbf{Static source}} & \multicolumn{2}{c}{\textbf{Moving source}} \\
    \cmidrule(lr){2-3} \cmidrule(lr){4-5}
    & \textbf{Solo} & \textbf{Swarm} & \textbf{Solo} & \textbf{Swarm} \\
    \midrule
    Time to $\mathbf{x}_0$ (median) & \SI{329.63}{\second} & \SI{59.16}{\second} & \SI{203.66}{\second} & \SI{52.90}{\second} \\
    Time to $\mathbf{x}_0$ (max) & \SI{584.67}{\second} & \SI{124.68}{\second} & \SI{359.30}{\second} & \SI{149.70}{\second} \\
    Tracking time (average) & \SI{167.01}{\second} & \SI{180}{\second} & \SI{44.50}{\second} & \SI{133.30}{\second} \\
    Tracking time (max) & \SI{180}{\second} & \SI{180}{\second} & \SI{122.43}{\second} & \SI{180}{\second} \\
    Estimation error (median) & \SI{2.59}{\meter} & \SI{2.26}{\meter} & \SI{20.34}{\meter} & \SI{6.70}{\meter} \\
    \bottomrule
  \end{tabular}
\end{table}



\section{EXPERIMENTS}
\label{sec:experiments}

A series of experiments was conducted to evaluate the localization capabilities in real-world conditions utilizing 3 \acp{MAV} built on the modular hardware platform \cite{hert2023hardware}.
The core software stack is the MRS UAV System \cite{baca2021mrs} with ROS\footnote{\url{https://www.ros.org/}} Noetic running on the Intel NUCi7 onboard computer.
The sensor payload consists of a MiniPIX TPX3 miniature Compton camera with a \SI{2}{\milli\meter} CdTe detector, a Holybro Pixhawk4 flight controller, a Garmin LIDAR-Lite v3 rangefinder used as altimeter, and a NEO-M8N \ac{GNSS} receiver (\reffig{fig:experiment_uavs}).

The radioactive sources used in the experiments are laboratory isotopes of Cesium-137 with activity\footnote{\SI{1}{\becquerel} = 1 nuclear decay event per second} \SI{179}{\mega\becquerel} and \SI{2}{\giga\becquerel}.
The position of the radiation source is chosen randomly for each experiment and the ground truth is measured using \ac{GNSS} or estimated from a bird's-eye view footage of the experiment.
The search is conducted in an open area \SI{40}{\meter}  $\times$ \SI{50}{\meter} in size with no obstacles, and only one radiation source is present at a time.

\begin{figure}[htb]
  \centering
  \includegraphics{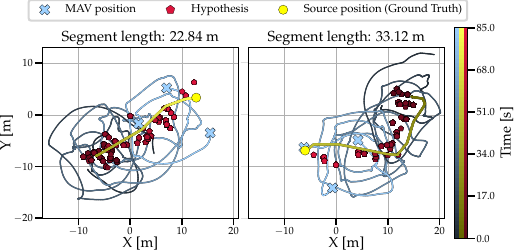}
  \caption{
    A top-down view of the moving radiation source experiments.
    The radiation source is carried by a remotely controlled ground robot.
    The estimated position of the source (hypothesis) is shown in red, and the path taken by the \acp{MAV} is shown in blue.
    The change in color brightness illustrates the time progression of the experiments.
  }
  \label{fig:experiment_tracking}
\end{figure}

For the experiments with a moving radiation source, the \SI{2}{\giga\becquerel} Cesium-137 sample was carried by a remotely controlled Boston Dynamics Spot (\reffig{fig:experiment_spot}).
The motion of the radiation source consisted of ad-hoc straight and curved segments taken at various speeds, and focused on finding the tracking limits in real-world conditions.
The moving source experiment was performed 7 times resulting in over \SI{60}{\minute} of accumulated flight data from each device.
The results of the experiments are summarized in \reftab{tab:experiments_table}.
Our approach was found to be capable of tracking the motion of the \SI{2}{\giga\becquerel} source up to a speed of \SI{3}{\meter\per\second} using 3 \acp{MAV}.
If using just 1 or 2 \acp{MAV}, the system was not able to localize and follow the source moving at this speed in any of the attempts we tried.
\reffig{fig:experiment_tracking} shows snapshots of the moving source experiments, highlighting the flight paths, estimated position of the radiation source, and the ground truth.

\begin{table}[htbp]
  \centering
  \caption{
    \label{tab:experiments_table}
    Performance evaluation of the real-world deployment using 3 \acp{MAV} and Cesium-137 as the radiation source.
  }
  \begin{tabular}{l cc|c}
    \toprule
    & \textbf{\SI{179}{\mega\becquerel}} & \textbf{\SI{2}{\giga\becquerel}} & \textbf{\SI{2}{\giga\becquerel} (moving)} \\
    \midrule
    Time to $\mathbf{x}_0$ (median) & \SI{105.46}{\second}    & \SI{29.55}{\second}   & \SI{45.88}{\second} \\
    Time to $\mathbf{x}_0$ (max)    & \SI{115.88}{\second}    & \SI{34.80}{\second}    & \SI{139.98}{\second} \\
    Estimation error (median)       & \SI{2.65}{\meter}       & \SI{3.49}{\meter}     & \SI{3.50}{\meter} \\
    Estimation error (average)      & \SI{5.04}{\meter}       & \SI{4.89}{\meter}     & \SI{6.24}{\meter} \\
    \bottomrule
  \end{tabular}
\end{table}

\begin{figure}[htbp]
  \centering

  \subfloat{
    \hspace{-0.35cm}
    \imagewithletter{\includegraphics[width=0.328\columnwidth]{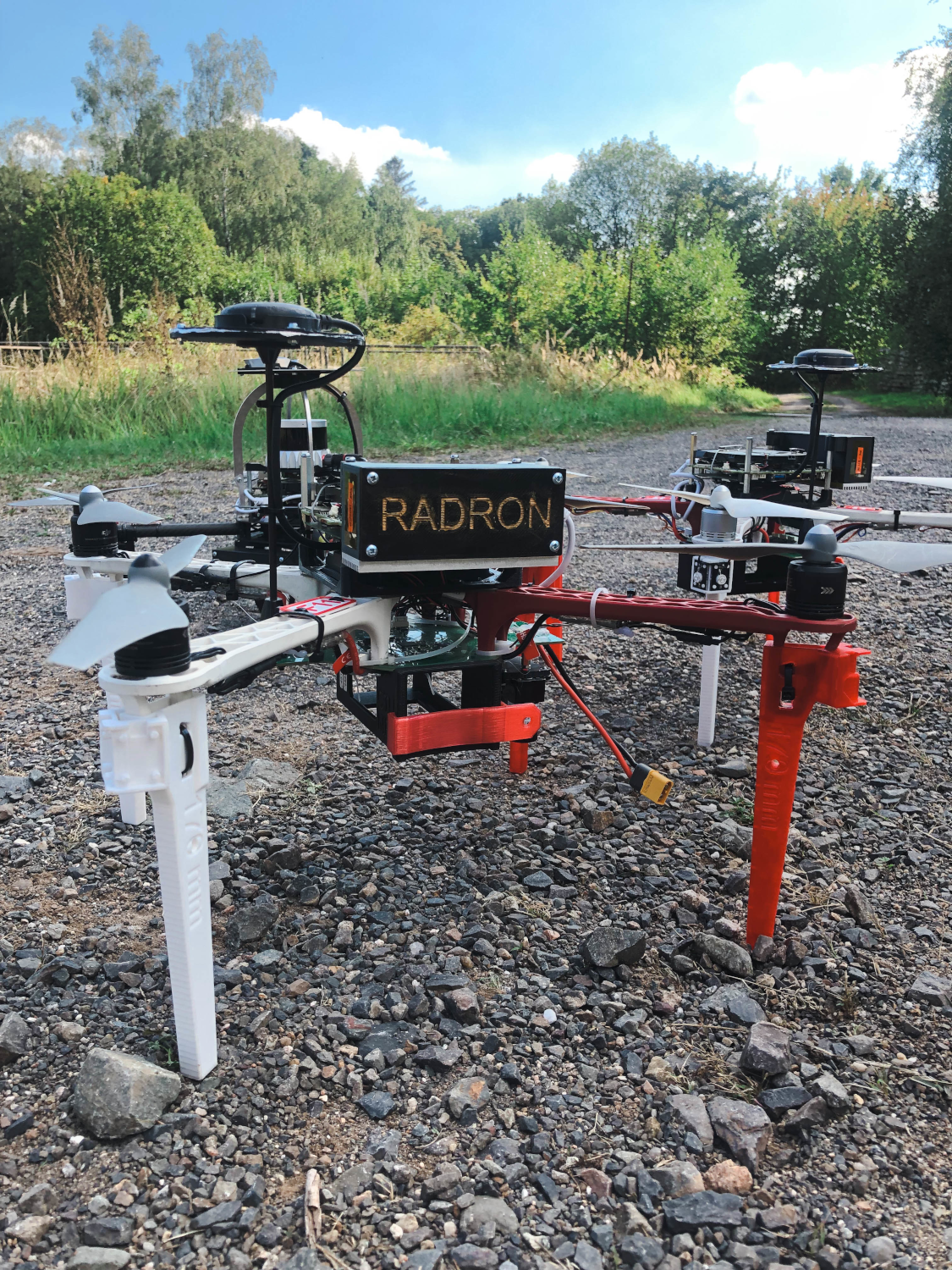}}{(a)}{fig:experiment_uavs}
  }
  \subfloat{
    \hspace{-0.55cm}
    \imagewithletter{\includegraphics[width=0.657\columnwidth]{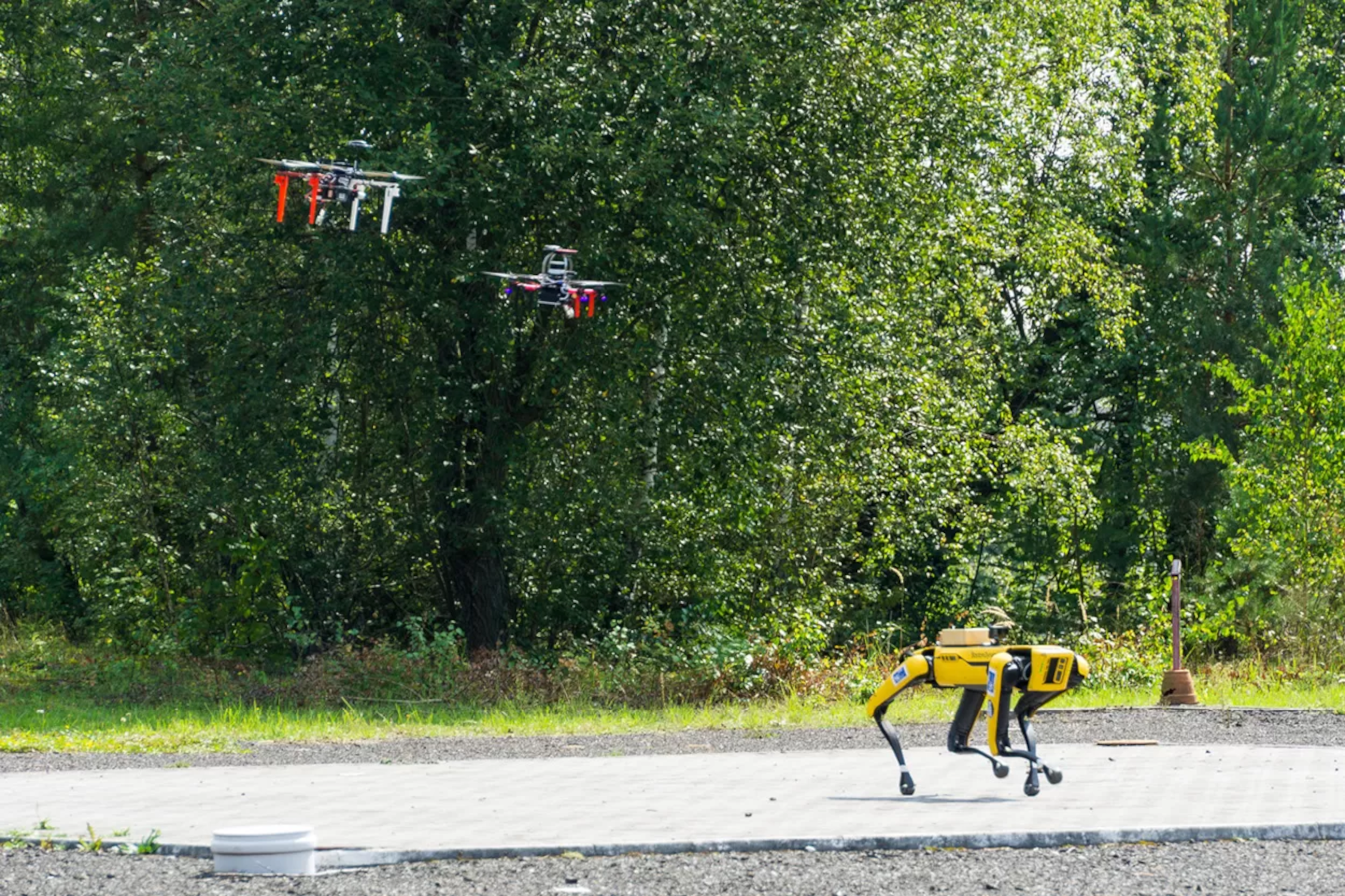}}{(b)}{fig:experiment_spot}
  }

  \caption{
    Photos taken during the real-world experiments showing the \ac{MAV} platforms \textbf{(a)} with the MiniPIX TPX3 Compton camera mounted on the front.
    To shield the sensitive electronics from dust, the Compton camera is sealed in a 3D-printed case (black) which gamma rays easily penetrate.
    For the mobile source tracking experiments \textbf{(b)}, the radiation source was carried by the Boston Dynamics Spot legged robot.
  }
  \label{fig:experiment_photos}
\end{figure}

\section{CONCLUSION}

  The simulation and experimental results conclusively demonstrate a clear advantage of a cooperating \ac{MAV} swarm over a single vehicle across all performance metrics for radiation source localization.
  The superiority is clearly rooted in the fusion of measurements taken from multiple viewpoints simultaneously, which results in significant speed up of the initialization phase and dramatic improvements in source localization accuracy.
  A truly unique aspect of our novel approach is the ability to localize and track a moving radiation source.
  The empirical results highlight the critical importance of the initialization phase, as a poor initial hypothesis may prevent the subsequent estimation and tracking method from converging due to an insufficient rate of radiation events reaching the Compton cameras.



\bibliographystyle{IEEEtran}
\bibliography{IEEEabrv,references}

\end{document}